\pdfoutput=1

\documentclass[11pt]{article}

\usepackage{acl}

\usepackage{times}
\usepackage{latexsym}
\usepackage{booktabs}
\usepackage{xcolor}

\usepackage[T1]{fontenc}
\usepackage{subcaption}

\usepackage[utf8]{inputenc}

\usepackage{microtype}
\usepackage{subfiles} 

\usepackage{algorithm}
\usepackage{algpseudocode}
\usepackage{graphicx}
\usepackage{multirow}
\usepackage{amsmath}
\usepackage{makecell}

\title{DEFT-UCS: Data Efficient Fine-Tuning for Pre-Trained Language Models via Unsupervised Core-Set Selection}

\author{Devleena Das\thanks{Contributed during an internship at Accenture Labs, SF} \\
  Georgia Institute of Technology \\
  \texttt{ddas41@gatech.edu} \\\And
  Vivek Khetan \\
  Accenture Labs \\
  \texttt{vivek.a.khetan@accenture.com} \\}

\begin{document}
\maketitle
\begin{abstract}
Recent advances have led to the availability of many pre-trained language models (PLMs); however, a question that remains is how much data is truly needed to fine-tune PLMs for downstream tasks? In this work, we introduce DEFT-UCS, a data-efficient fine-tuning framework that leverages unsupervised core-set selection to identify a smaller, representative dataset that reduces the data needed to fine-tune PLMs for the text-generation task of text-editing. We examine the efficacy of DEFT-UCS across multiple text-editing tasks, and compare to the state-of-the art text-editing model, CoEDIT. Our results demonstrate that DEFT-UCS models are just as accurate as CoEDIT, across eight different datasets consisting of six different editing tasks, while finetuned on 70\% less data.

\end{abstract}

\section{Introduction}
How much data do we need to fine-tune a pre-trained language model (PLM) for a specific downstream task? While successes in language modelling have led to numerous publicly available PLMs and ability to produce fine-tuned models for downstream tasks - the answer mostly remains, ``as large as possible, and of good quality''. For example,  Alpaca, an instruction-following model, is trained with 52k data samples \cite{taori2023alpaca}. Similarly, CoPoet, a collaborative poetry writing system is fine-tuned using 87k data samples \cite{chakrabarty2022help}. MetaMath, a math-reasoning LLM is fine-tuned with 395k data samples \cite{yu2023metamath}. Although fine-tuning PLMs on specific task results in performance gain, acquiring large amounts of data for fine-tuning is not easy for real-world applications which often require niche knowledge and domain expertise. 

Researchers have explored variety of methods primarily focused on improving the computational efficiency of fine-tuning, including parameter-efficient fine-tuning approaches (PEFT) to reduce computational costs by optimizing parameter updates \cite{fu2023effectiveness, hu2021lora} as well as leveraging active-learning for iteratively selecting data samples during training \cite{su2022selective, diao2023active}. Instead, our work focuses on improving the \textit{data efficiency} of PLM fine-tuning without requiring iterative fine-tuning. Specifically, we explore how to fine-tune PLMs with significantly less data samples and without a cost to model performance. Related to language models, researchers have experimented with different core-set selection metrics \cite{paul2021deep, sorscher2022beyond} to improve the data efficiency during \textit{pre-training}. \citet{marion2023less} demonstrated how perplexity, L2-Error Norm (EL2N) and memorization can be utilized to select smaller, good quality datasets for pre-training. Similarly, \cite{attendu2023nlu} leverage EL2N to dynamically remove data samples with high EL2N between training epochs. However, these metrics assume access to task data and reference models to perform dataset pruning. In real world applications, utilizing such supervised, data-pruning metrics are less realistic since large amounts of annotated task-specific data may be costly to acquire. This leads us to our main research question:
\textit{How can we leverage unsupervised data pruning to fine-tune PLMs for downstream tasks in a more data efficient manner?}

In this work, we introduce a new data-efficient fine-tuning framework, DEFT-UCS, that uses unsupervised core-set selection to minimize the amount of labelled data needed to fine-tune PLMs for the text-generation task of text-editing. Our framework is inspired by  \cite{sorscher2022beyond}, who 
utilize clustering-based dataset pruning to reduce training samples for image-classification models, and to the best of our knowledge, our framework is the first to leverage
unsupervised core-set selection for data-efficient fine-tuning of PLMs.
  
We investigate the utility of DEFT-UCS in fine-tuning PLMs for text-generation across eight different datasets consisting of six different text-editing tasks, and compare DEFT-UCS models to the state-of-the-art text-editing model, CoEDIT\cite{raheja2023coedit}. Our contributions are as follows: 
\begin{itemize}
    \item We introduce DEFT-UCS, a data-efficient-fine tuning framework that leverages unsupervised core-set selection via clustering to identify a smaller representative set of data needed to fine-tune PLMs. 
    
    \item We show that DEFT-UCS, utilizing only 32.5\% of CoEDIT's training data, is able to produce fine-tuned models with improved accuracy on four different text-editing tasks, and similar accuracy on two text-editing tasks compared to CoEDIT \cite{raheja2023coedit}.
    
    \item We performed a human evaluation with 3 evaluators to assess the quality of text-edits from our DEFT-UCS model. Evaluators found edits generated by DEFT-UCS model as similar or preferred over CoEDIT \cite{raheja2023coedit}.

\end{itemize}

\section{Related Works}
\paragraph{\textbf{Efficient Fine-Tuning of LLMs}} 
Most work on efficient fine-tuning techniques for LLMs have primarily focused on parameter-efficient fine-tuning (PEFT) approaches \cite{fu2023effectiveness, hu2021lora}, improving computation efficiency by updating a subset of model parameters. Recently, there has been an increasing focus on improving the data-efficiency of LLMs, considering how to pre-train and fine-tune LLMs with smaller subsets of data \cite{zhou2023lima,mukherjee2023orca,chen2023skill,marion2023less, attendu2023nlu,ivison2022data}. For instance, \citet{zhou2023lima} introduce LIMA, an approach to fine-tune LLaMA \cite{touvron2023llama} with only 1k diverse and high quality samples. However, the LIMA approach is underspecificed without a general subsampling procedure. Also, \citet{chen2023skill} develop Skill-It!, which creates efficient datasets by learning hierarchical relationships between samples. However, identifying hierarchical relationships is non-trivial and not all datasets may include them. 
More closely related to our work, \citet{ivison2022data} leverage K-Nearest Neighbors to learn multiple data-efficient fine-tuned models for individual tasks. Instead, we aim to learn a single data-efficient fine-tuned model that performs competitively across a variety of datasets. Similarly, \citet{marion2023less} utilize perplexity and EL2N, to find smaller datasets for LLM pre-training, and \citet{attendu2023nlu} uses EL2N to iteratively remove unimportant samples during fine-tuning. Both \citet{marion2023less} and \citet{attendu2023nlu} assume access to task data to train various PLMs for few epochs used to calculate EL2N and perplexity. In contrast, we leverage unsupervised core-set selection, omitting the need for any reference model during the dataset sampling step.

\paragraph{\textbf{Core-Set Selection \& Dataset Distillation}}
Several works in ML have developed variety of 
core-set selection \cite{har2005smaller} and dataset pruning \cite{paul2021deep} methods to find smaller subsets of data needed to train deep learning models without model performance loss. CRAIG \cite{mirzasoleiman2020coresets} finds core-sets by approximating gradient calculations, while RETRIEVE \cite{killamsetty2021retrieve} finds core-sets by optimizing for model loss. Also, \citet{yang2022dataset} utilize Influence Functions \cite{koh2017understanding} to prune redundant samples. A unifying idea among these methods is the need for labelled data. 

Alternatively, core-set selection methods for unlabelled data have used clustering methods. \citet{birodkar2019semantic} use Agglomerative clustering to find semantic similarities among data points and prune redundant samples. Similarly, \citet{sorscher2022beyond} use vanilla k-means clustering and distances to cluster centroids for pruning easy and hard samples. Recently, data distillation algorithms have also been developed to improve data-efficient model training \cite{zhou2023dataset}. Typically, data distillation methods generate new synthetic datasets in which data samples are edited to preserve more information for performance generalization \cite{lei2023comprehensive}. Our work considers efficient core-set selection without the generation of a synthetic dataset. Specifically, our work builds upon the unsupervised clustering approach in \citet{sorscher2022beyond} applied to computer vision tasks, to fine-tune PLMs in a data-efficient manner.

\paragraph{Instruction Tuning for Text-Editing}
Recently, Instruction tuning of PLMs has shown impressive success in its ability to enable PLMs to follow instructions as well as improvement in generalization across various tasks in zero/few shot settings \cite{min-etal-2022-metaicl, Wei2021FinetunedLM}. Training models to explicitly follow natural language instructions has become increasingly popular for text-editing tasks as well. \citet{shu2023rewritelm} develop RewriteLM by fine-tuning PaLM \cite{chowdhery2022palm} variants for the task of rewriting long-form texts. Similarly, \citet{schick2022peer} develop PEER by fine-tuning T5 \cite{raffel2020exploring} variants to emulate the collaborative writing process. Additionally, \citet{raheja2023coedit} develop CoEDIT by fine-tuning Flan T5 \cite{chung2022scaling} models to perform single and compositional edits across multiple edit tasks. Furthermore, \citet{zhang2023multi} produce an instruction-tuned LLaMA model that improve text-editing capabilities. A commonality across these works include the usage of large scale datasets for fine-tuning. For example, CoEDIT \cite{raheja2023coedit} and \citet{zhang2023multi} leverage datasets with 82k and 60k examples, respectively. In our work, DEFT-UCS maximizes model performance of fine-tuned models in a \textit{data efficient} manner by finding a representative, smaller dataset needed for fine-tuning. We investigate the efficacy of our DEFT-UCS framework to instruction fine-tune PLMs for eight text-editing tasks.

\section{Problem Formulation}
\label{sec:problem-formulation}
We formulate DEFT-UCS as an unsupervised core-set selection problem \cite{sorscher2022beyond} in contrast to existing dataset pruning methods which primarily use supervised core-set selection \cite{attendu2023nlu, marion2023less}. 

Specifically, let $D$ represent an existing large dataset,  $P$ represent a PLM, and $M_{D}$ represent $P$ fine-tuned on $D$. Our DEFT-UCS framework aims to find a representative core-set $D_{c} \subset D$ such that leveraging $D_{c}$ can fine-tune $P$ and result in a fine-tuned model $M_{D_{c}}$ with comparable performance to $M_{D}$. Note, we refer to comparable evaluation performance in the form of both quantitative NLP metrics and qualitative human evaluations. Specific to unsupervised core-set selection, DEFT-UCS finds $D_{c}$ without needing $D$ to include annotations or labels. Thus, we find $D_{c}$ by only using the input samples $\{x_{1}..x_{n}\}$ within $D$. These input samples, in the context of instruction fine-tuning, represent task instructions and input texts. 

To perform unsupervised core-set selection, we build upon the SoTA clustering-based core-set selection method by \citet{sorscher2022beyond}, given its extensive evaluations against other supervised-based core-set selection methods. While \citet{sorscher2022beyond} demonstrate the efficacy of clustering-based core-set selection for ImageNet \cite{deng2009imagenet}, our work is the first to investigate the effectiveness of clustering-based core-set selection in non-classification tasks, such as fine-tuning PLMs for multiple text-editing tasks.

\begin{algorithm} 
\caption{Unsupervised Core-set Selection (UCS)}\label{alg:cap}
\textbf{Input:}{~$D_{remain} = \{x_0, x_1 ... x_{n}\}$ - Large Dataset}\\
\textbf{Input:}{~$K$ - Num. of Clusters } \\ 
\textbf{Input:}{~$A$ - Amount of samples per cluster} \\
\textbf{Input:}{~$\alpha$, ~$\beta$, - Sampling Weights}\\
\textbf{Output:}{~$D_{c} = \{x_j.. x_{p}\}$ - Core-Set} \\ 
\begin{algorithmic} [1]
\State $D_{c} = \emptyset$
\State $D_{embed}$ = ComputeEmbedding($D_{remain}$)
\State $Cl_{1:K}, Ce_{1:K}$ = KMeans($D_{embed}$, $K$)
\For{$i$ in $K$}
\For{$d$ in ${Cl}_{i}$}
\State $dist_{list}$ = StoreCosineDistance($d$, $Ce_{i}$) 
\EndFor
\State $dist_{sorted}$ = sort($dist_{list}$)
\State $D_{sampled}$ = $dist_{sorted}$[0 : $\alpha$*A] \par 
\hskip\algorithmicindent + $dist_{sorted}$[ -$\beta$*A:]   
\State $D_{c}$ = updateCoreSet($D_{sampled}$, $D_{c}$)
\EndFor \\
\Return  $D_{c}$ 
\end{algorithmic}
\end{algorithm}

\begin{figure}[t!]
\centering
  \includegraphics[width=0.42\textwidth]{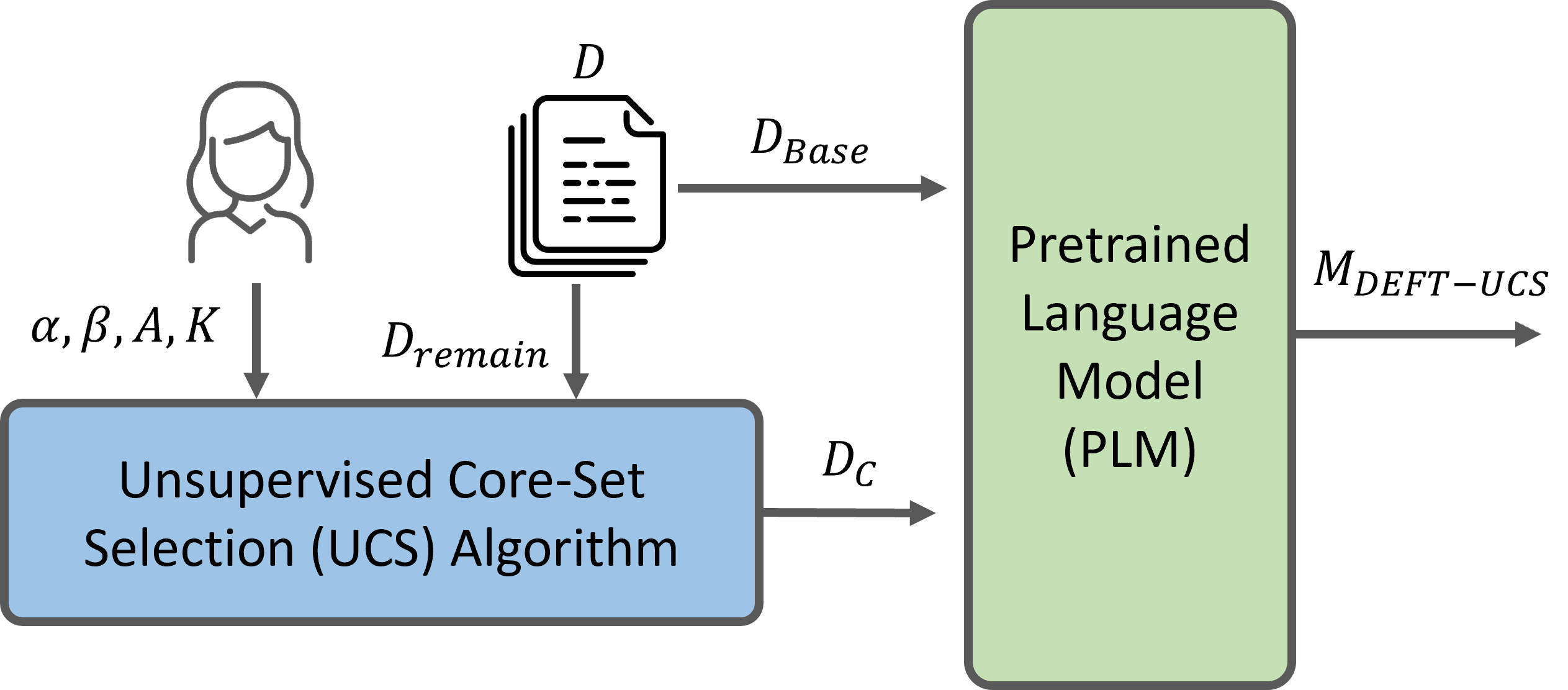}
  \caption{Our DEFT-UCS framework utilizes unsupervised core-set selection (UCS) to find a core-set of data $D_{c}$, as well as  initial seed data, $D_{base}$ to produce a fine-tuned PLM, $M_{DEFT-UCS}$. }
  \label{fig:DEFT-Coedit}
\end{figure}

\section{DEFT-UCS Framework}
\label{sec:method}
Figure \ref{fig:DEFT-Coedit} outlines our DEFT-UCS framework which leverages \textit{unsupervised, clustering-based core-set selection} (\textit{UCS}) to find a subset of $D$ that fine-tunes a PLM without compromising model performance. We consider a scenario in which there exists an initial amount of data, $D_{base} \subset D$, that is sampled in a stratified manner to provide an overall representation of the downstream fine-tuning task. Let $D_{remain}$ represent the remaining data after $D_{base}$ is sampled. The goal of UCS is to then find a core-set $D_{c} \subset D_{remain}$ that enriches $D_{base}$ such that $D_{c}$ and $D_{base}$, together, form a representative subset that can be used to fine-tune a PLM and result in a fine-tuned model $M_{DEFT-UCS}$ with comparable performance to $M_{D}$, a PLM fine-tuned with $D$. In Algorithm 1, we detail the crux of our DEFT-UCS
framework, the UCS method.

\subsection{Clustering in UCS} The first step in UCS includes transforming $D_{remain}$ into a meaningful embedding representation $D_{embed}$. UCS clusters $D$ based on its latent-space representation, using previously learned embedding spaces, such as sentenceBert \cite{reimers2019sentence}. Choosing an appropriate embedding representation is important, given that such representation impacts the downstream clustering task within UCS. In Section \ref{sec:DEFT-on-CoEDIT}, we detail the types of learned embedding spaces we evaluate and the best embedding representation found for encoding sentence-based datasets.

Given $D_{embed}$, we perform K-Means clustering to separate $D_{embed}$ into $K$ clusters. Note, the value of $K$ is dependent on $D$, and defining $K$ requires domain knowledge about the dataset to understand the different categories or tasks represented in $D$. Alternatively, $K$ can be automatically derived using metrics such as Silhouette Score \cite{shahapure2020cluster}. The resulting $K$ clusters, $Cl_{1:K}$, and cluster centroids, $Ce_{1:K}$, are utilized to compute the cosine distance between each data sample $d$ in a cluster $Cl_{i}$, and corresponding centroid $Ce_{i}$. 

\subsection{Sampling $\boldsymbol{D_{c}}$ in UCS} We leverage the clustering categorization presented in \citet{sorscher2022beyond} to sample $D_{c}$ from $D_{remain}$. Specifically, \citet{sorscher2022beyond} explain that data samples can be categorized as ``easy'' or ``hard'' examples. In the context of unsupervised clustering, \citet{sorscher2022beyond} leverage a data sample's distance to its cluster centroid to define easy and hard samples. Therefore, easy/hard samples within a cluster are those closest/furthest to the cluster centroid. Given such definition, in UCS, we retrieve a weighted sampling of easy and hard samples from each cluster, denoted as $D_{sampled}$. The $\alpha$ and $\beta$ weights control the distribution of easy and hard samples in $D_{sampled}$, and $A$ represents the total number of samples retrieved per cluster. 

Note, $D_{base}$, $K$, $A$, $\alpha$, and $\beta$ are hyperparameters within DEFT-UCS, manually set by domain-experts. Given this is the first work, to our knowledge, to propose data-efficient fine-tuning for PLMs leveraging UCS, we perform an exhaustive investigation on how these hyperparameters influence fine-tuning performance (see Section \ref{sec:all-results}). Future work includes investigating  automatic selection of such hyperparameters.

\section{DEFT-UCS Applied to Text-Editing}
\label{sec:DEFT-on-CoEDIT}
We evaluate the utility of DEFT-UCS in the context of instruction-based fine-tuning for multiple text editing tasks. To our knowledge, the current SoTA instruction fine-tuned text-editing LM is CoEDIT ($M_{CoEDIT}$)\footnote{https://github.com/vipulraheja/coedit} trained on dataset $D_{CoEDIT}$ \cite{raheja2023coedit}. Overall, $D_{CoEDIT}$ 
includes 82k good-quality edit instructions spanning six different edit-tasks \cite{raheja2023coedit} ($D_{CoEDIT}$ detailed in Appendix \ref{sec:appendix-coedit-dataset}). Given the data quality in $D_{CoEDIT}$ and SoTA performance of $M_{CoEDIT}$, we apply DEFT-UCS to $D_{CoEDIT}$. Below, we detail the hyper-parameter choices in DEFT-UCS in the context of $D_{CoEDIT}$.

\subsection{$\boldsymbol{D_{Base}}$ in CoEDIT} Recall $D_{Base}$ refers to initial data sampled in a stratified manner used for fine-tuning. In our work, stratified sampling is performed based on the different tasks represented in $D$. During our evaluations, we study how the size of $D_{Base}$ may influence hyperparameter selection within our UCS algorithm for producing a well-performing $M_{DEFT-UCS}$. In the context of CoEDIT, we experiment with $D_{Base}=\{10\%, 20\%,..80\%\}$, representing 10\% to 80\% of $D_{CoEDIT}$. Note, $D_{CoEDIT}$ is a fully annotated dataset; however, when performing core-set selection $D_{c} \subset D$, we only consider the input sentences.

\subsection{DEFT-UCS Hyperparameters} Given that $D_{CoEDIT}$ includes seven edit-intentions, we set $K=7$, allowing the K-Means Clustering within UCS to separate $D_{remain}$ into 7 clusters. Additionally, recall from Sec. \ref{sec:method} that $\alpha$ and $\beta$ represent the sampling weights for extracting easy and hard data samples from each cluster to form $D_{sampled}$. To understand the upper and lower bound effects of $\alpha$ and $\beta$, we study three variants of $D_{sampled}$, representing three different sampling types: $D^{hard}_{sampled}$, $D^{easy}_{sampled}$ and $D^{rand}_{sampled}$. Specifically, $D^{hard}_{sampled}$ is represented by $\alpha=0$ and $\beta=1.0$, $D^{easy}_{sampled}$ is represented by $\alpha=1.0$ and $\beta=0$, and $D^{rand}_{sampled}$ approximates $\alpha=0.5$ and $\beta=0.5$, denoting random samples extracted per cluster. We also experiment with sampling different amounts of data from each cluster, denoted by $A=\{285, 570, 857\}$. Such settings of $A$ approximate  
$\{2000, 4000, 6000\}$ total samples from $D_{remain}$ respectively, and represent $\{2.5\%, 5\%, 7.5\%\}$ percent of $D_{remain}$.

\subsection{Dataset Embedding} Recall that the UCS algorithm in DEFT-UCS performs clustering using a learned embedding representation of the input data samples. We investigate several embedding representations and select the best embedding representation by its ability to inform accurate clusters.  Specifically, we study sentence-level encodings from Sentence-T5 \cite{ni2021sentence}, BART \cite{lewis2019bart} CLS token embeddings, as well as averaged word token embeddings from Flan-T5 \cite{chung2022scaling}. From an ablation study, our results demonstrate that leveraging Sentence-T5 \cite{ni2021sentence} results in the best K-Means Clustering performance. The ablation study results are in Appendix \ref{sec:appendix-embedding-representation}.

\subsection{Model Fine-Tuning} \citet{raheja2023coedit}
develop CoEDIT-Large, CoEDIT-xl, and CoEDIT-xxl by fine-tuning Flan-T5's Large, XL and XXL models, respectively. In our work, we focus our comparisons against CoEDIT-Large, referred to as $M_{CoEDIT}$. Therefore, in our framework, we fine-tune Flan-T5-Large, producing $M^{Flan-T5-LG}_{DEFT-UCS}$. Details on our fine-tuning implementation are in Appendix \ref{sec:appendix-finetuning-details}.

\begin{center}
\begin{table}[t]
\resizebox{0.47\textwidth}{!}{%
\begin{tabular}{ l | l }\toprule
 \textbf{Evaluation Dataset} & \textbf{Edit Task} \\
 \midrule
 TurkCorpus \cite{xu2016optimizing} & Simplification
  \\ 
 Asset \cite{alva2020asset} & Simplification
 \\ 
  Iterator Coherence \cite{du2022understanding} & Coherence 
  \\ 
 Iterator Clarity \cite{du2022understanding} & Clarity  
 \\ 
  Iterator Fluency \cite{du2022understanding} & Fluency  
  \\ 
 Iterator Global \cite{du2022understanding} & Clarity, Coherence, Fluency  
 \\ 
 JFLEG \cite{napoles2017jfleg} & Grammar Correction  
 \\ 
  WNC \cite{pryzant2020automatically} & Neutralization  
  \\ 
 \bottomrule
\end{tabular}}
\caption{A list of datasets, spanning six editing tasks, on which we evaluate our DEFT-UCS models.}
\label{tab:eval_datasets}
\end{table}
\end{center}


\vspace{-0.6cm}
\section{Experiments}

\subsection{Evaluation Datasets}
Table \ref{tab:eval_datasets} presents eight test datasets used in our evaluation. We performed evaluations across six different edit tasks including simplification, coherence, clarity, fluency, grammar correction and neutralization improvement. See Appendix \ref{sec:appendix-eval-datasets} for dataset details. For fair comparisons, these datasets include the publicly available datasets evaluated by CoEDIT \cite{raheja2023coedit}, and are present in  several text-editing benchmarks, including EDITEVAL \cite{dwivedi2022editeval}.

\subsection{Metrics} 
We examine SARI \cite{xu-etal-2016-optimizing} and ROUGE-L \cite{lin2004rouge} scores for our quantitative evaluations. SARI scores are also utilized in prior text-editing tasks \cite{raheja2023coedit}.
During our human evaluation, we analyze users' perceived accuracy percentage (\textit{PA\%}), which measures the percent of times users select a text-editing model for producing accurately edited sentences. 

\subsection{Baselines}
We compare our fine-tuned models via DEFT-UCS, $M_{DEFT-UCS}$, to the following baselines.

\paragraph{CoEDIT-Large} The primary baseline of our work is the original CoEDIT-Large model \cite{raheja2023coedit}, $M_{CoEDIT}$, which uses the entire 82k samples in $D_{CoEDIT}$ to fine-tune Flan-T5 Large. To compare against $M_{CoEDIT}$, we utilize the released CoEDIT model\footnote{https://huggingface.co/grammarly/coedit-large} and compare SARI and ROUGE-L scores for each evaluation dataset.

\paragraph{LIMA Approach} 
We also compare our DEFT-UCS method to the LIMA approach \cite{zhou2023lima}.
Following the LIMA approach of using high quality and diverse 1k data points, we select 1k data samples via stratified random sampling from $D_{CoEDIT}$ for fine-tuning Flan-T5. We refer to such LIMA-inspired model as $M_{LIMA}$. Prior work by Raheja et al. \cite{raheja2023coedit} validate the high-quality data samples in $D_{CoEDIT}$, and stratified random sampling ensures data diversity, allowing all editing tasks within $D_{CoEDIT}$ to be equally represented.

\paragraph{Non-Instruction Fine-Tuned LLMs} We also compare our $M_{DEFT-UCS}$ with LLamA2-7B ($M_{LLAMA2-7B}$) \cite{touvron2023llama}, Flan-T5-Large ($M_{FLAN-T5-LG}$) \cite{chung2022scaling} and BLOOM-560M ($M_{BLOOM-560M}$) \cite{scao2022bloom}, in Zero-Shot settings, to understand how $M_{DEFT-UCS}$ compares to non-instruction fine-tuned LLMs.


\begin{figure*}[t]
\centering
\begin{subfigure}[b]{0.49\textwidth}
  \centering
  \includegraphics[width=\textwidth]{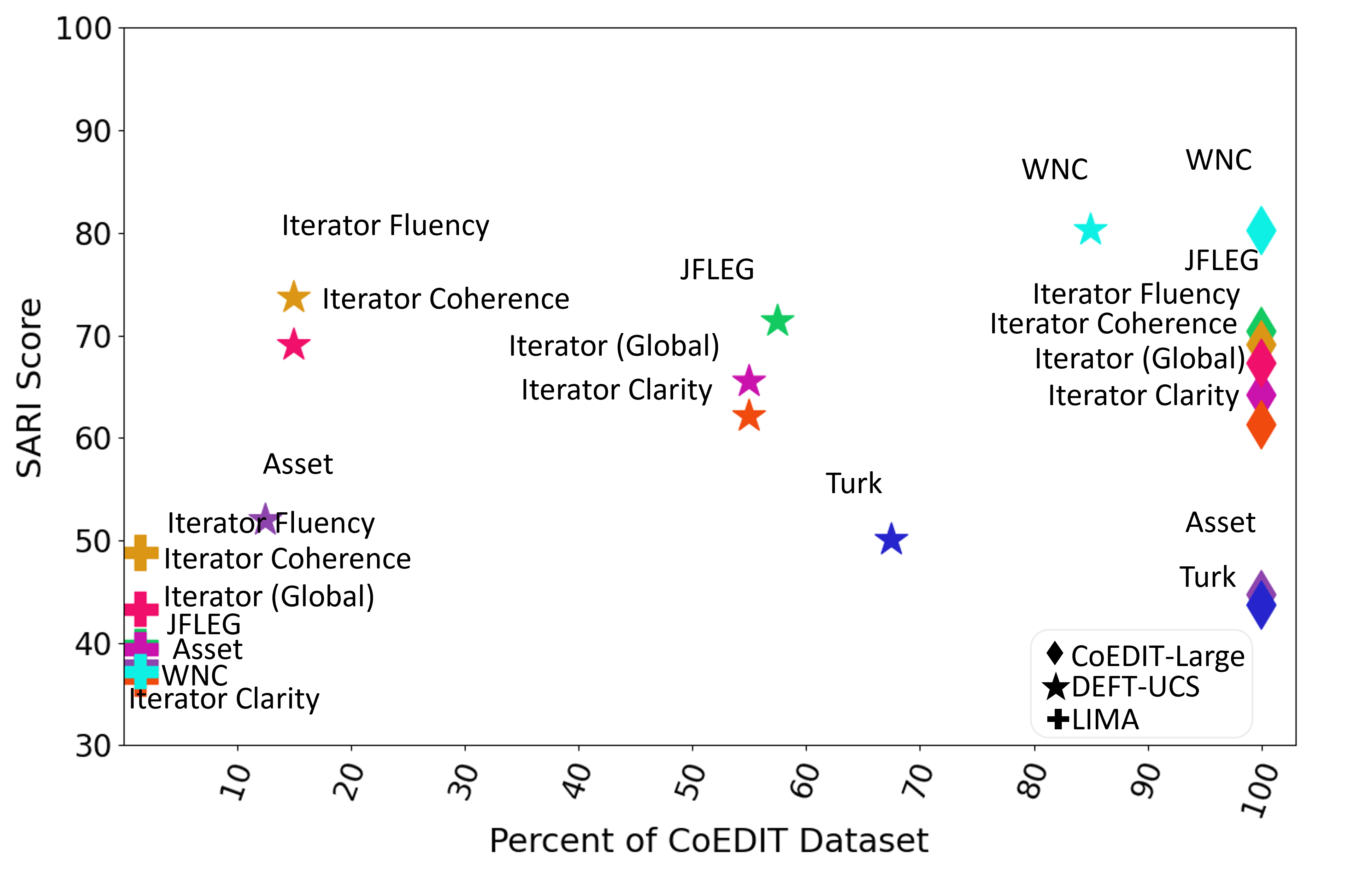}
  \caption[]%
  {{}}
\end{subfigure}\quad
\begin{subfigure}[b]{0.485\textwidth}
  \centering
  \includegraphics[width=\textwidth]{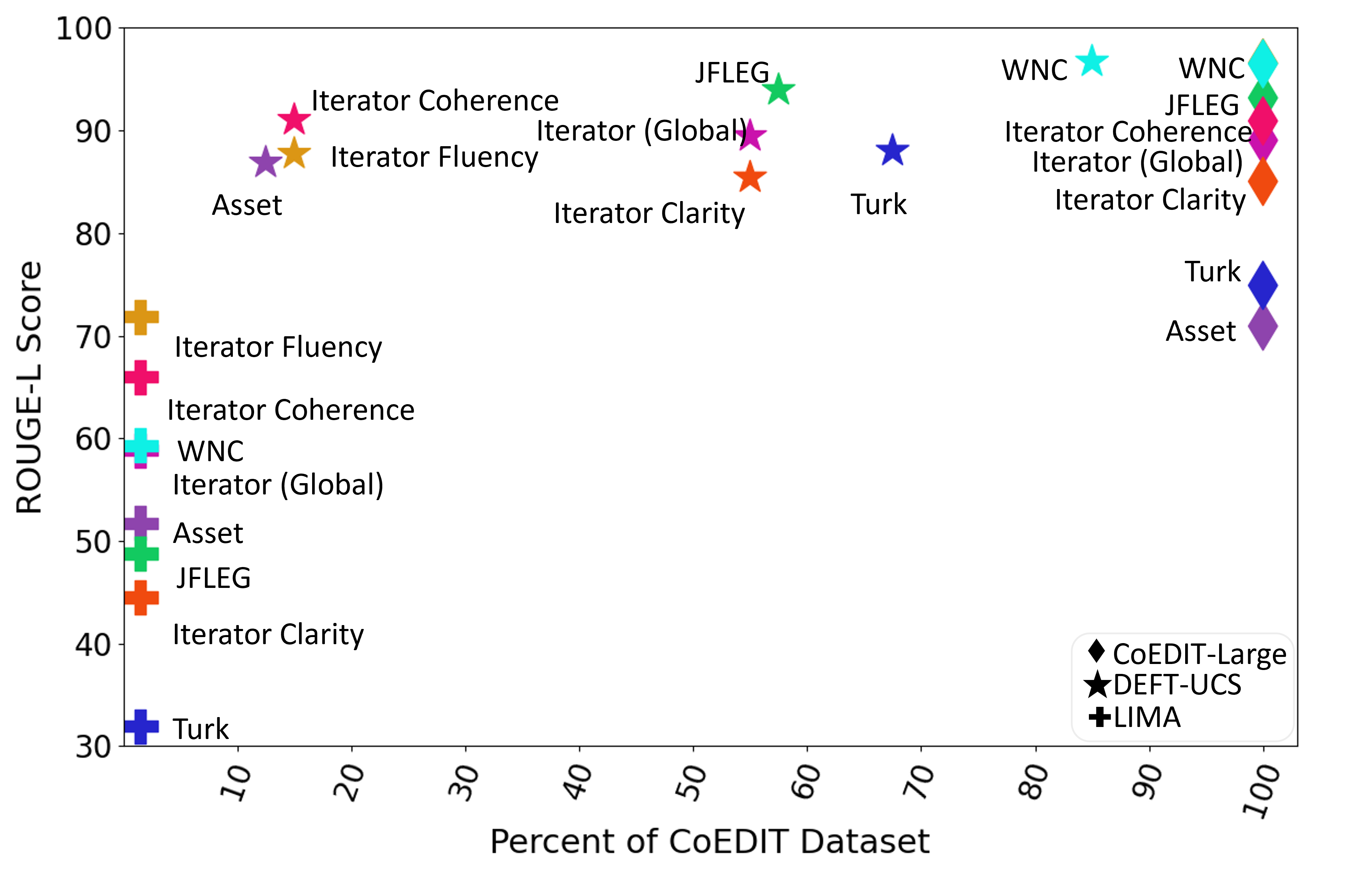}
    \caption[]%
     {{}}
     \end{subfigure}
\caption{Comparisons between the CoEDIT model \cite{raheja2023coedit}, LIMA-inspired model $M_{LIMA}$ \cite{zhou2023lima}, and our DEFT-UCS models with respect to SARI (a) and ROUGE-L (b) scores.}
\label{fig:rq1}
\end{figure*}

\begin{table*}[h]
\resizebox{\textwidth}{!}{%
\renewcommand{\arraystretch}{1.4} 
\begin{tabular}{ p{2.5cm} | ccccccccc }\toprule
       \textbf{Models} & \textbf{Turk} & \textbf{Asset} &\textbf{Iterator Coherence} & \textbf{Iterator Clarity} & \textbf{Iterator Fluency} & \textbf{Iterator Global} & \textbf{JFLEG} & \textbf{WNC} \\
    \midrule
        $M^{Flan-T5-LG}_{DEFT-UCS}$ & \textbf{46.6} / \textbf{81.1} & \textbf{46.8} / \textbf{76.9} & \textbf{68.9} / 90.9 & \textbf{61.8} / \textbf{85.3} & \textbf{69.9} / \textbf{96.9}  & \textbf{64.7} / \textbf{89.1} & 70.2 / 93.1 & 79.0 / 96.5\\
        $M_{CoEDIT}$  & 43.7 / 74.9 & 44.7 / 70.9 & 67.3 / \textbf{91.1} & 61.3 / 85.1 & 69.1 / 96.6 & 64.2 / 89.0 & \textbf{70.4} / \textbf{93.2} & \textbf{80.2} / \textbf{96.5}\\
        \midrule
        $M_{LIMA}$ & 23.8 / 31.9 & 37.8 / 51.7 & 43.3 / 65.9 & 36.5 / 55.5 & 48.8 / 71.9  & 39.4 / 58.9 & 39.7 / 48.8 & 37.2 / 59.3\\
        $M_{LLAMA2-7B}$ & 36.8 / 17.3 & 41.6 / 20.3 & 35.8 / 26.2 & 41.2 / 28.5 & 40.4/ 33.8 & 38.3/ 29.7 & 46.0 / 17.0 & 27.3 / 17.2\\
        $M_{FlAN-T5-LG}$ & 32.3 / 59.1 & 41.3 / 74.7 & 36.7 / 52.4 & 34.3 / 54.3 & 37.9 / 64.9 & 35.5 / 57.7 & 51.3 / 80.9 & 30.7 / 48.9\\
        $M_{BLOOM-560M}$ & 27.3 / 7.7 & 32.0 / 8.2 & 19.1 / 8.8 & 20.6 / 9.7 & 16.3/ 8.2  & 19.6 / 9.5 & 27.9 / 4.9 & 18.8/ 8.1\\
    \bottomrule
\end{tabular}}
\caption{Comparisons between the overall best DEFT-UCS model, $M^{FLan-T5-LG}_{DEFT-UCS}$ with all other baselines, with the first value representing SARI score and second value representing ROUGE-L score. Note, scores for LLAMA-7B and BLOOM-560 model (Zero-shot) generations are calculated by first removing the prepended input sequence.}
\label{tab:best_deft_baselines}
\end{table*}

\begin{figure*}[t]
\centering
\begin{subfigure}[b]{0.49\textwidth}
  \centering
  \includegraphics[width=\textwidth]{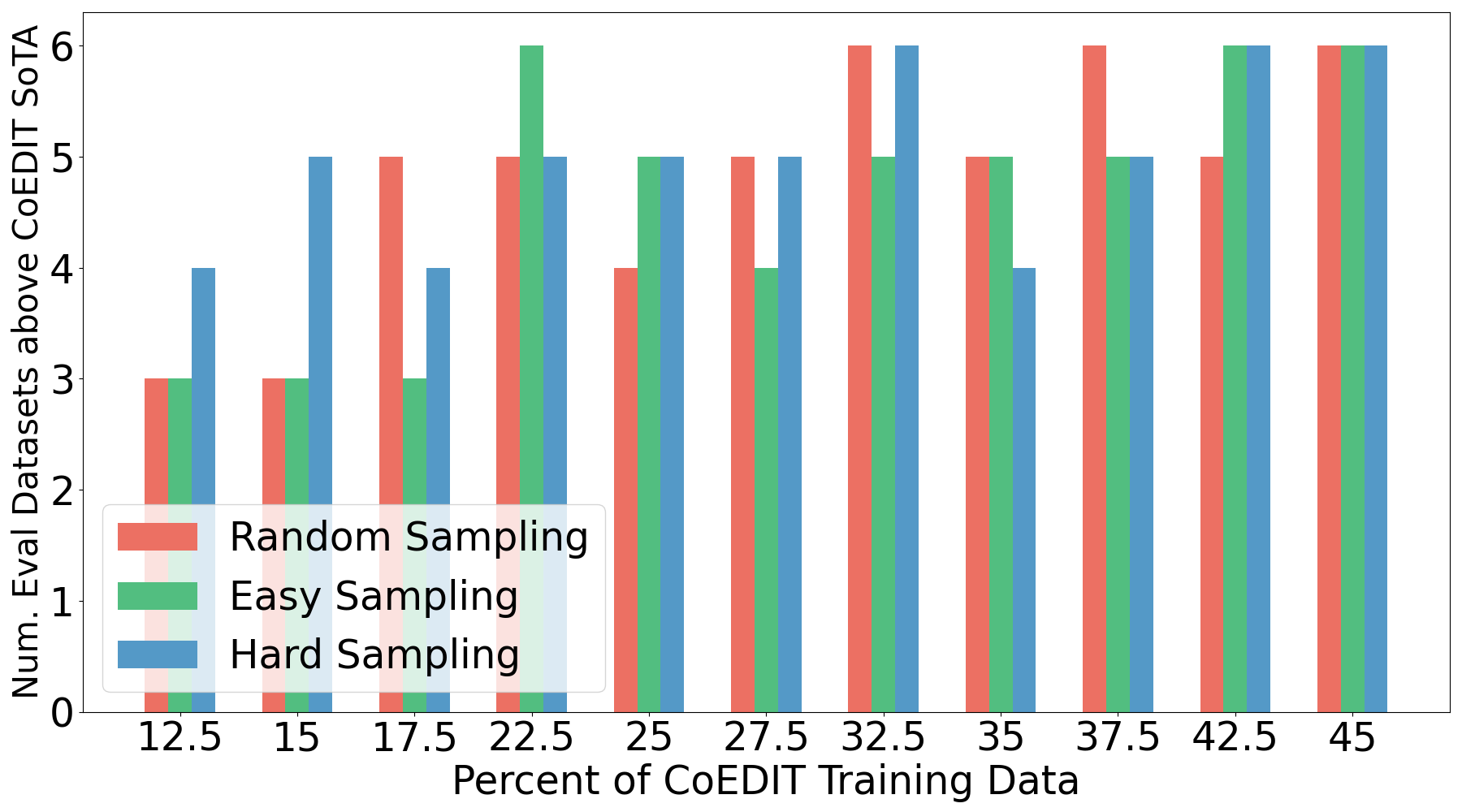}
  \caption[]%
  {{}}
\end{subfigure}\quad
\begin{subfigure}[b]{0.485\textwidth}
  \centering
  \includegraphics[width=\textwidth]{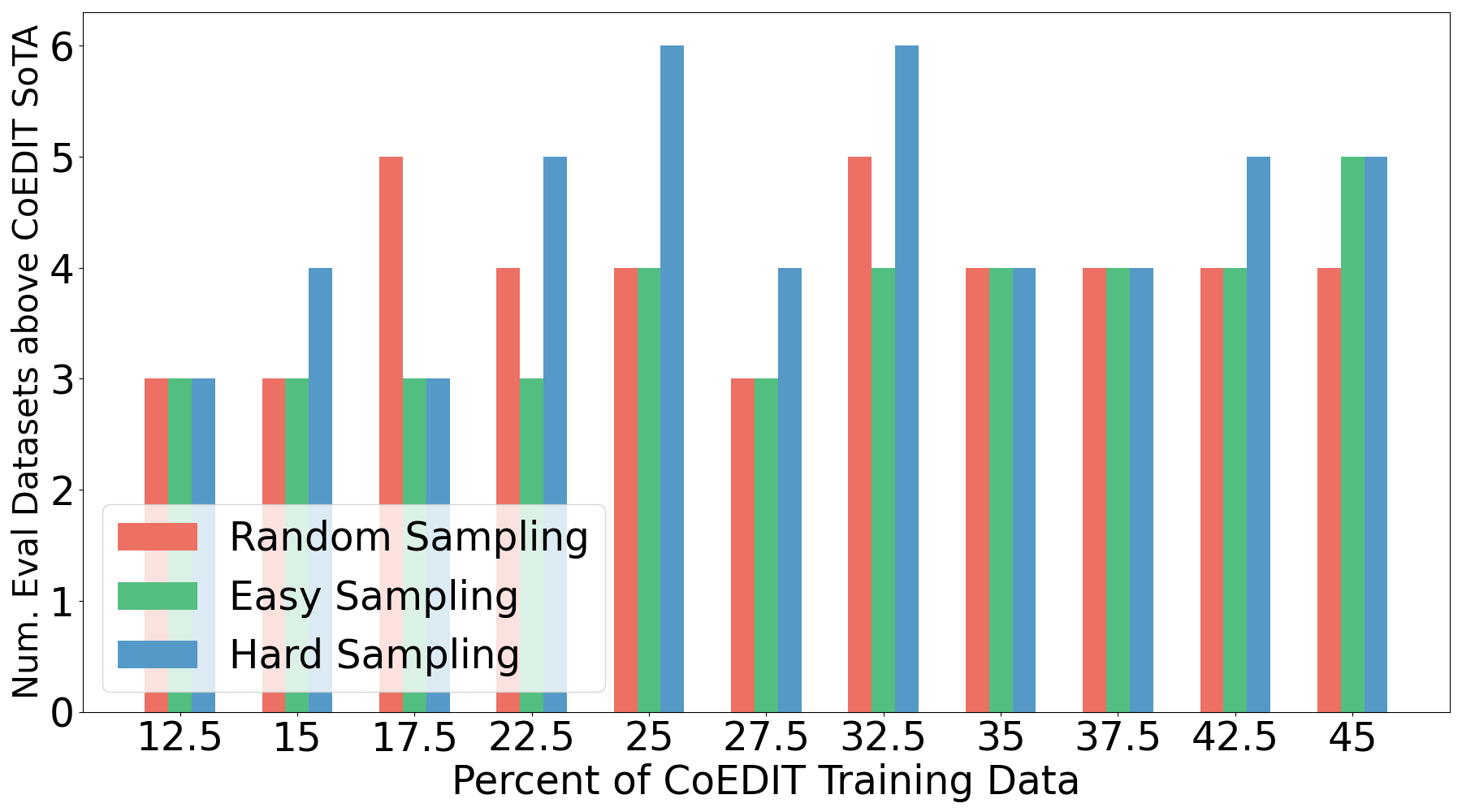}
    \caption[]%
     {{}}
     \end{subfigure}
\caption{Utilizing hard sampling in UCS results in a best, overall DEFT-UCS model that requires only 32.5\% of $D_{CoEDIT}$ to beat 6/8 evaluation datasets considering SARI (a) and ROUGE-L (b) scores.}
\label{fig:best_deft}
\end{figure*}

\begin{figure*}[t]
\centering
\begin{subfigure}[b]{1.\textwidth}
  \centering
  \includegraphics[width=\textwidth]{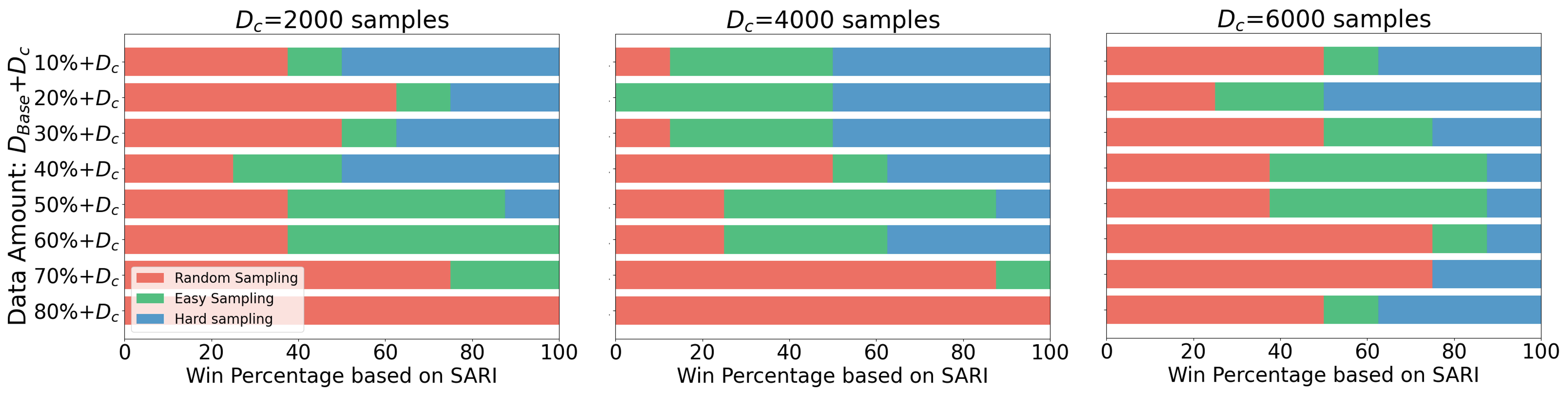}
  \caption[]%
  {{}}
\end{subfigure}\quad
\begin{subfigure}[b]{1.\textwidth}
  \centering
  \includegraphics[width=\textwidth]{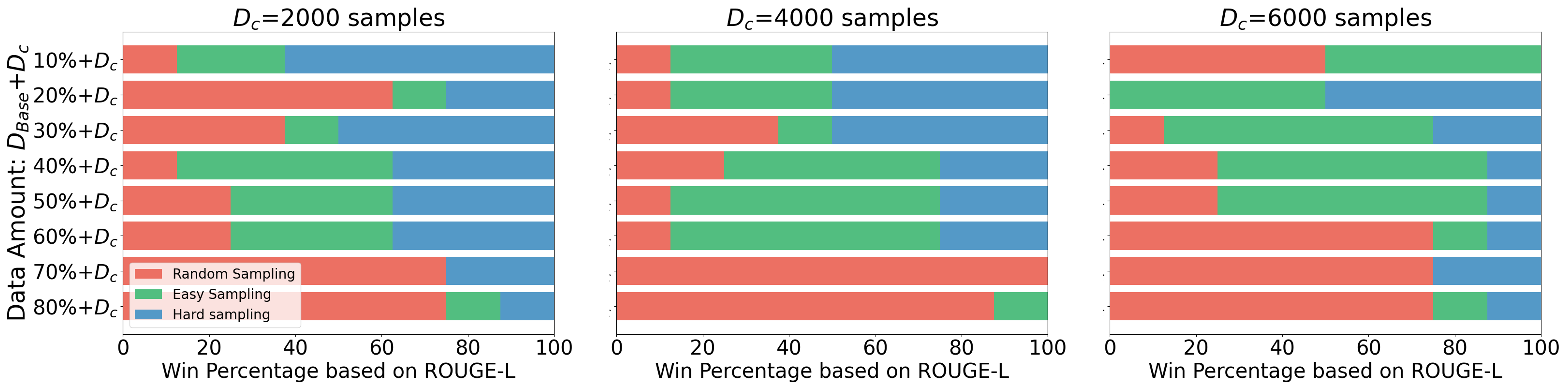}
    \caption[]%
     {{}}
     \end{subfigure}
\caption{With less $D_{base}$, leveraging hard sampling in our DEFT-UCS leads to better performing models (winning \%); as $D_{base}$  increases, random sampling leads to better performing models. }
\label{fig:d_base_influence}
\end{figure*}

\section{Results}
\label{sec:all-results}
Our results below show that DEFT-UCS can provide a \textit{data-efficient} method for producing competitive fine-tuned models for six different text-editing tasks.

\label{sec:deft-vs-coedit}
\subsection{DEFT-UCS vs. CoEDIT}
Figure \ref{fig:rq1} shows that our DEFT-UCS framework generates fine-tuned models with comparable performance to $M_{CoEDIT}$ in terms of SARI (Fig. \ref{fig:rq1}a) and Rouge-L (Fig. \ref{fig:rq1}b) scores, using lower fractions of $D_{CoEDIT}$. These results indicate that unsupervised core-set selection within DEFT-UCS can effectively find a $D_{c}$ for fine-tuning without compromising downstream task performance.

The DEFT-UCS models in Figure \ref{fig:rq1} reflect the existence of \textit{a} competitive DEFT-UCS model, and depending on the evaluated text-editing task, a different fraction of $D_{CoEDIT}$ results in the \textit{most} competitive performances. For example, to achieve comparable performance on the WNC dataset for the neutralization task, a DEFT-UCS model needs above 80\% of $D_{CoEDIT}$. In contrast, for the Asset dataset and simplification task, around 12\% of  $D_{CoEDIT}$ is needed to surpass $M_{CoEDIT}$ SARI and ROUGE-L scores. We hypothesize that subjectivity in the neutralization task (WNC) increases the complexity of the data samples and more data is required to fine-tune a competitive model in comparison to less subjective editing tasks such as, text-simplification (Asset). 

\subsection{DEFT-UCS vs. LIMA Approach} We observe across all evaluation tasks, $M_{LIMA}$ has lower SARI and ROUGE-L scores compared to $M_{CoEDIT}$ and our DEFT-UCS models. These results show that the LIMA \cite{zhou2023lima} approach may not be generalizable to domain-specific LM tasks such as text-editing and more experimentation is needed to understand its limitations. Moreover, these results indicate that smarter sampling techniques that go beyond data quality and diversity are needed for competitive model performances, such as considering distance metrics in embedding spaces as utilized in DEFT-UCS.
 
\subsection{Overall DEFT-UCS Model}
\label{sec:best-deft-results}
In Section 7.1, we found that the most competitive DEFT-UCS model for each evaluation dataset uses a different fraction of $D_{CoEDIT}$. Therefore, we performed an additional analysis to study which combination of hyper-parameters result in an 
overall best-performing DEFT-UCS model, one that achieves or surpasses $M_{CoEDIT}$ performances on most evaluation datasets using a much smaller fraction of $D_{CoEDIT}$.
Fig. \ref{fig:best_deft}(a) and Fig. \ref{fig:best_deft}(b) show that fine-tuning Flan-T5 Large with only 32.5\% of $D_{CoEDIT}$ and performing hard sampling ($\alpha=0$, $\beta=1.0$), results in the best overall DEFT-UCS model, $M^{FLAN-T5-LG}_{DEFT-UCS}$, surpassing $M_{CoEDIT}$ SARI and ROUGE-L scores on six of the eight evaluation datasets. Overall, 32.5\% represents the smallest fraction of $D_{CoEDIT}$ that results in competitive SARI and ROUGE-L scores on most evaluation datasets. 

Note, 32.5\% of $D_{CoEDIT}$ is composed of $D_{base}$, initial data available for fine-tuning, and $D_{c}$, the output of UCS within DEFT-UCS. In the context of $M^{FLAN-T5-LG}_{DEFT-UCS}$, $D_{base}$ is a stratified 30\% subset from $D_{CoEDIT}$, and $D_{c}$ is composed of another 2.5\% of $D_{remain}$ ($A=2000$ samples per cluster) retrieved from UCS by performing hard sampling.  

\paragraph{Model Performance}
Table \ref{tab:best_deft_baselines} shows the SARI and ROUGE-L scores of our best DEFT-UCS model, $M^{FLAN-T5-LG}_{DEFT-UCS}$, fine-tuned with only 32.5\% of $D_{CoEDIT}$. 
We find that $M^{FLAN-T5}_{DEF-UCS}$ performs better than 
$M_{LIMA}$ and $M_{FLAN-T5-LG}$ on four datasets, and  comparably on two datasets, WNC and JFLEG.  These results emphasize that a much smaller fraction of $D_{CoEDIT}$ can be used to produce a comparable fine-tuned text-editing model. We also observe that $M_{LLAMA2-7B}$ and $M_{BLOOM-560}$ have much lower ROUGE-L scores compared to all other models. After examining model generated outputs, we see that lower ROUGE-L scores are attributed to long, repeated sentences from $M_{LLAMA2-7B}$ $M_{BLOOM-560}$. Appendix \ref{sec:appendix-qualitative-examples} provides example edited sentences from each model. 

\paragraph{Influence of $\boldsymbol{D_{Base}}$ \& Sampling Methods}
Based on downstream tasks, the amount of $D_{Base}$ may vary. Thus, we analyze how the size of $D_{base}$ may influence the sampling method utilized in DEFT-UCS for producing best-performing models. Figure \ref{fig:d_base_influence} summarizes the win percentages among the three sampling methods (random sampling, easy sampling, hard sampling) as the size of $D_{base}$ increases. Win percentage is defined as the percent of times a particular sampling method achieves the highest SARI (Fig. \ref{fig:d_base_influence}a) or ROUGE-L (Fig. \ref{fig:d_base_influence}b) score across all evaluation datasets. From Figure \ref{fig:d_base_influence}a and Figure \ref{fig:d_base_influence}b, we observe that as $D_{Base}$ increases, even across different $D_{c}$ amounts, random sampling results in better SARI and ROUGE-L performances compared to easy and hard sampling. However, with lower amounts of $D_{Base}$, hard sampling results in better performance. We hypothesize that with lower amounts of $D_{Base}$, sampling harder examples may allow the model to generalize to unseen examples. Interactions between $D_{Base}$ and sampling type may be dataset and task dependent, and future work should experiment with these hypotheses for different task-specific applications. 

\begin{center}
\begin{table}[t]
\resizebox{0.47\textwidth}{!}{%
\begin{tabular}{ l | l }\toprule
 \textbf{Model} & \textbf{Perceived Accuracy (\textit{PA\%})} \\
 \midrule
 $M^{Flan-T5-LG}_{DEFT-UCS}$ & 83.8 \%
  \\ 
 $M_{CoEDIT}$ \cite{raheja2023coedit} & 70.5\%
 \\ 
 \bottomrule
\end{tabular}}
\caption{Perceived accuracy from  human evaluation.}
\label{tab:human-eval}
\end{table}
\end{center}

\vspace{-0.7cm}
\subsection{Human Evaluation}
We hired three computer scientists with English as their primary language for our human evaluation. We created a human-eval test set by randomly sampling 35 examples from seven text-editing dataset in Table \ref{tab:eval_datasets}.\footnote{We did not sample from Iterator Global since such dataset is a combination of Iterator Clarity, Fluency and Coherence.} For each sample in the human-eval test set,  evaluators were provided two edited sentence generated using $M^{FLAN-T5-LG}_{DEFT-UCS}$ and $M_{CoEDIT}$. Evaluators were then asked to select the most accurately edited sentence.
Given that many edited sentences from $M^{FLAN-T5-LG}_{DEFT-UCS}$ and $M_{CoEDIT}$ were similar or identical, evaluators were able to select more than one edited-sentence as accurately edited. To reduce bias, the generated sentence ordering from the models was randomized. 

Table \ref{tab:human-eval} summarizes the average perceived accuracy percentages (\textit{PA\%}). Overall, our $M^{FLAN-T5-LG}_{DEFT-UCS}$ results in higher PA\% compared to $M_{CoEDIT}$. We also calculated the inter-rater reliability score to understand the agreement among evaluators on their PA\%, and found moderate agreement with a Fleiss-Kappa \cite{fleiss1973equivalence} score of 0.44. These results indicate that evaluators perceived our $M^{FLAN-T5-LG}_{DEFT-UCS}$ to produce accurately edited-sentences with comparable quality between $M_{CoEDIT}$ and $M^{FLAN-T5-LG}_{DEFT-UCS}$.

\section{Conclusion}


We introduce DEFT-UCS, a data-efficient fine-tuning framework that leverages unsupervised core-set selection to find the minimum amount of data needed to fine-tune a PLM for text-editing tasks. Our best performing DEFT-UCS model, fine-tuned with only 32.5\% of the CoEDIT dataset \cite{raheja2023coedit},  perform comparably to the SoTA text-editing model CoEDIT \cite{raheja2023coedit}, and superior to the LIMA approach \cite{zhou2023lima} both in quantitative and qualitative evaluations. Human evaluators also preferred edits generated by DEFF-UCS model over CoEDIT. 

These results show the overall utility of our DEFT-UCS framework towards data-efficient fine-tuning of PMLs in text-editing tasks. To better understand the generalizability of our framework, we plan to first apply it to a variety of text-generation tasks in the future. Subsequently, we aim to benchmark the efficacy of different data-sampling strategies across various PLMs for these tasks.

\paragraph{Limitations} The hyper-parameters within the UCS algorithm  of our DEFT-UCS framework are selected manually using task specific knowledge. Future work should consider how to automate the selection of these hyper-parameters. Additionally, while our UCS algorithm leverages the distance between data samples and centroid distance for defining sampling methods within DEFT-UCS, future work should explore other sampling methods informative to NLP tasks. Additionally, we show the utility of DEFT-UCS in the context of six different text-editing tasks; benchmarking the utility of DEFT-UCS in other task specific domains is needed to understand the scope of our framework. Similarly, more work is required to investigate the utility of DEFT-UCS in fine-tuning different PLMs for downstream NLP tasks. Future work also entails comparing the benefit of utilizing DEFT-UCS against PEFT \cite{fu2023effectiveness, hu2021lora} approaches, understanding whether DEFT-UCS in conjunction with PEFT can further improve the fine-tuning efficiency of PLMs.

\paragraph{{Ethics Statement}} We utilize a publicly available dataset from CoEDIT\footnote{https://huggingface.co/datasets/grammarly/coedit}. The dataset primarily focuses on non-meaning changing text edits and does not raise any privacy concerns. Nevertheless, the underlying autoregressive models may hallucinate and propagate biases. Before deploying for real world applications, considerations on how to incorporate user feedback for continual system improvement should be studied. Additionally, we have acknowledged the limitations of our DEFT-UCS framework and the need for more extensive benchmarking with various other PLMs and downstream tasks. Our work provides a initial set of results
and is an effort to motivate further research in data-efficient fine-tuning of PLMs.



\bibliography{custom}
\bibliographystyle{acl_natbib}

\clearpage
\appendix
\newpage

\section{DEFT-UCS Applied to CoEDIT}

\subsection{CoEDIT Dataset Details}
\label{sec:appendix-coedit-dataset}
The CoEDIT dataset, $D_{coEDIT}$, from \citet{raheja2023coedit} is comprised of several edit tasks, including fluency, coherence, clarity, paraphrasing, neutralization and formalization. As mentioned in \cite{raheja2023coedit}, the 82k data samples follow the format of $\langle instruction: source, target\rangle$ pairs. The source and target pairs come from a variety of different datasets related to each editing task. Table \ref{tab:coedit-training-dataset} summarizes the datasets utilized to represent each edit task in $D_{CoEDIT}$. The \textit{instruction} component are task-specific and generated from a pool of instructional prompts. For example, for a grammar correction task, an instruction could be ``Fix grammar errors'' or ``Fix grammatical errors in this sentence''. The list of all instructional prompts utilized are detailed in \cite{raheja2023coedit}. 

\subsection{DEFT-UCS Model Fine-Tuning Details}
\label{sec:appendix-finetuning-details}
Recall that all DEFT-UCS models in this paper are produced by fine-tuning Flan-T5 Large \cite{chung2022scaling}. We fine-tune Flan-T5 Large such that we can make accurate comparisons with $M_{CoEDIT}$\cite{raheja2023coedit} which represents a fine-tuned Flan-T5-Large model on $D_{CoEDIT}$. Furthermore, to remove any difference in model performances due to differing hyperparameters, we utilize the hyperparameters listed in \citet{raheja2023coedit}. Specifically, we use the Adam optimizer with a learning rate of 1e-4. All DEFT-UCS models in the main paper are trained for 5 epochs with early stopping and the model checkpoints with the best validation loss are saved. To perform fine-tuning, we leverage 4 A10G GPUs, from AWS G5 instances, using Deepspeed \cite{rasley2020deepspeed}, and the maximum source and target sequence length is set to 256. 

\begin{table}
\centering
\resizebox{0.45\textwidth}{!}{%
\begin{tabular}{ p{4cm} | p{5cm}}\toprule
 \textbf{Edit Task} & \textbf{Datasets in $D_{coEDIT}$} \\
 \midrule
 Fluency & NUCLE-14\newline Lang-8\newline BEA-19
  \\ \hline
 Coherence & DiscoFuse
   \\ \hline
Clarity\newline (Simplification) &  NEWSELA\newline
WikiLarge\newline
WikiAuto\newline
ParabankV2\newline
Iterator-Clarity
  \\ \hline
 Paraphrasing & ParabankV2
 \\ \hline
 Formalization & GYAFC 
  \\ \hline
Neutralization & WNC
 \\ 
 \bottomrule
\end{tabular}}
\caption{Data in $D_{CoEDIT}$\cite{raheja2023coedit} is comprised of samples from the above datasets. This table is a simplified version of Table 1 in \citet{raheja2023coedit}.}
\label{tab:coedit-training-dataset}
\end{table}
\section{Embedding Representations in UCS}
\label{sec:appendix-embedding-representation}
\subsection{Representation Details}
For K-means clustering to learn informative clusters, selecting the right latent space representation for the input data is important. In our application, an accurate embedding representation should allow each cluster to predominantly represent a certain type of editing task. For example all data related to editing for fluency should be clustered together, whereas all data related to grammar correction should be clustered together. To ultimately select an accurate embedding representation, we experimented with three different representations: sentence-level encoding from Sentence-T5 \cite{ni2021sentence}, BART CLS token embedding, as well as an averaged word token embedding from Flan-T5. As a brief summary, Sentence-T5 \cite{ni2021sentence} maps sentences to a 768 dimensional vector space using only the encoder from T5. Specifically, \citet{ni2021sentence} demonstrate that Sentence-T5 embeddings are able to lead to high performance in sentence transfer tasks. Similarly, we also experiment with BART \cite{lewis2019bart} CLS token embeddings, inspired by the notion that CLS token can provide informative representations of the input sentence for downstream tasks \cite{devlin2018bert}. We also experiment with an average pooling method of averaging all word embeddings of an input sequence, using the Flan-T5 model, to reach a sentence-level embedding.

\subsection{Representation Analysis}
\begin{figure*}[t]
\centering
\begin{subfigure}[b]{0.4\textwidth}
  \centering
  \includegraphics[width=\textwidth]{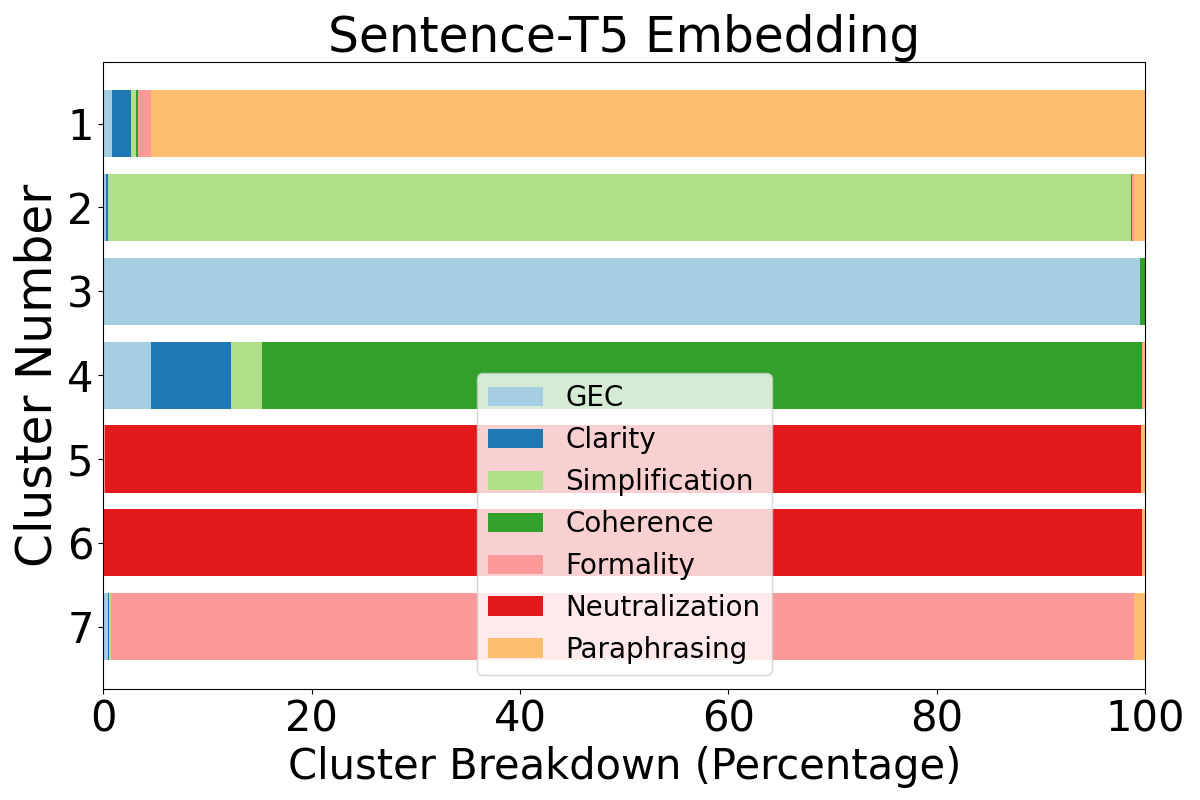}
  \caption[]%
  {{}}
\end{subfigure}\quad
\begin{subfigure}[b]{0.4\textwidth}
  \centering
  \includegraphics[width=\textwidth]{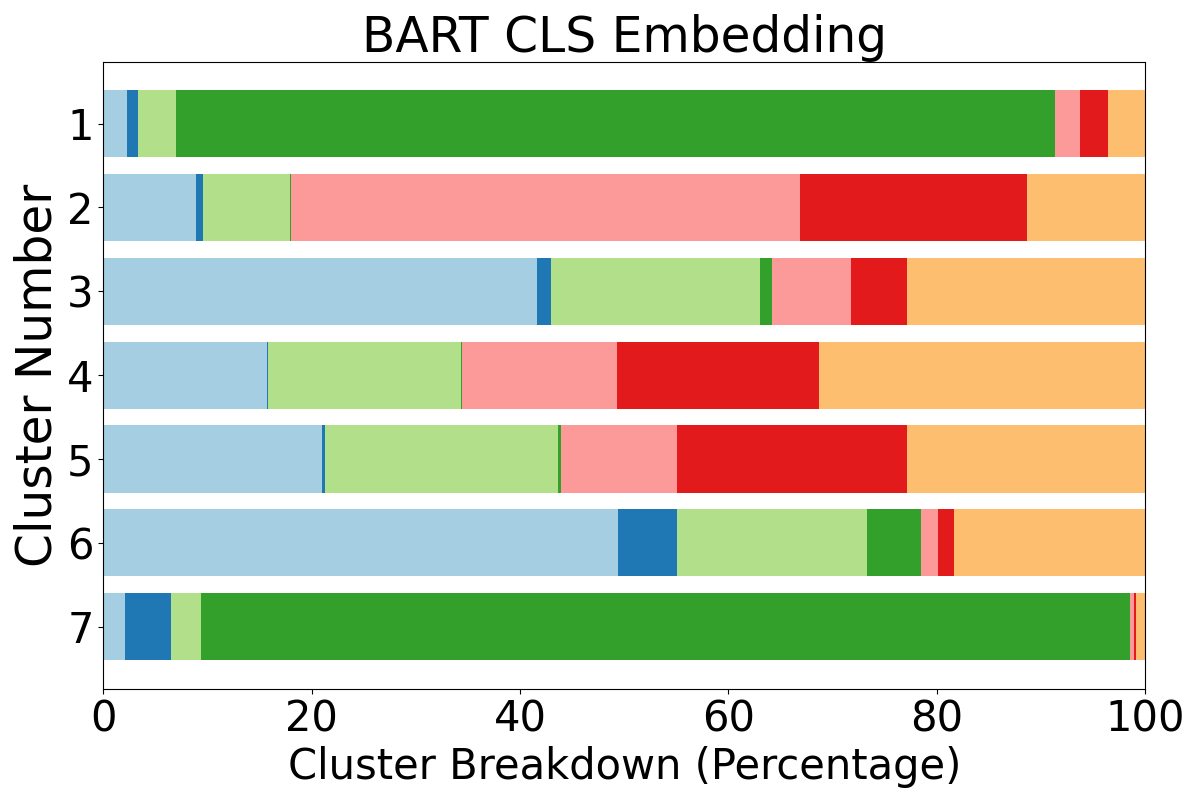}
    \caption[]%
     {{}}
     \end{subfigure}
     \begin{subfigure}[b]{0.4\textwidth}
  \centering
  \includegraphics[width=\textwidth]{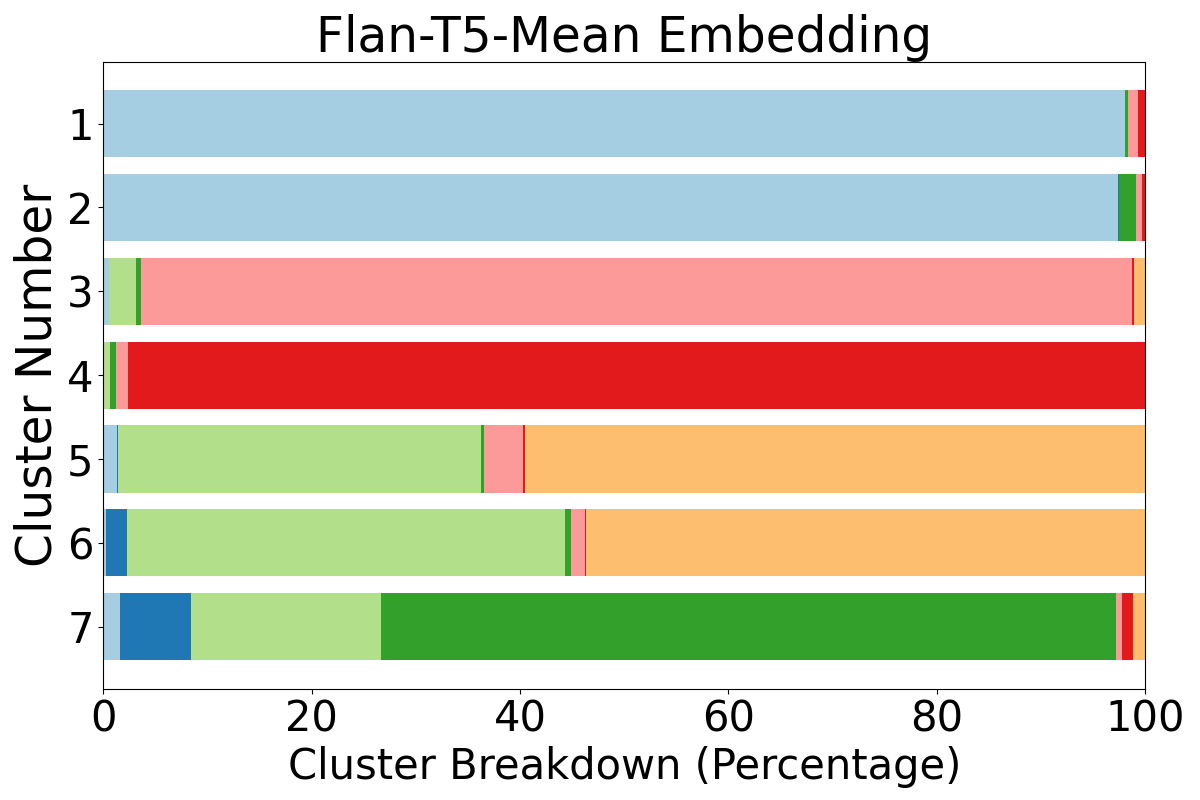}
    \caption[]%
     {{}}
     \end{subfigure}
\caption{Comparing the distribution of task-related data among clusters after performing K-Means when utilizing Sentence-T5 embedding (a), BART CLS embeddings (b) and averaged Flan-T5 word embeddings (c) for sentence representations.}
\label{fig:embedding-representation-analysis}
\end{figure*}

Figure \ref{fig:embedding-representation-analysis} demonstrates the K-means clustering results for each sentence-level embedding representation. Overall, we find that Sentence-T5 provides the strongest sentence-level embedding that allows the clustering algorithm to best separate input data based on its related editing task. Specifically, when analyzing Figure \ref{fig:embedding-representation-analysis}(a), we see that each cluster is largely comprised of a single edit-task. For example, cluster 1 largely includes data related to ``paraphrasing'', while cluster 4 largely includes data related to improving ``coherence''. In Figure \ref{fig:embedding-representation-analysis}(b) and Figure \ref{fig:embedding-representation-analysis}(c) we observe that the task specific data is more distributed among several clusters, indicating weaker cluster separation among the different editing task related data. Although the clusters formed via Sentence-T5 embeddings \cite{ni2021sentence} are not perfect, they offer the strongest separation of task-related data compared to the other embedding representations. Given these results, we leverage Sentence-T5 as our latent space representation when performing UCS.

\section{Evaluation Dataset Details}
\label{sec:appendix-eval-datasets}
\begin{figure*}[h!]
\centering
  \includegraphics[width=0.7\textwidth]{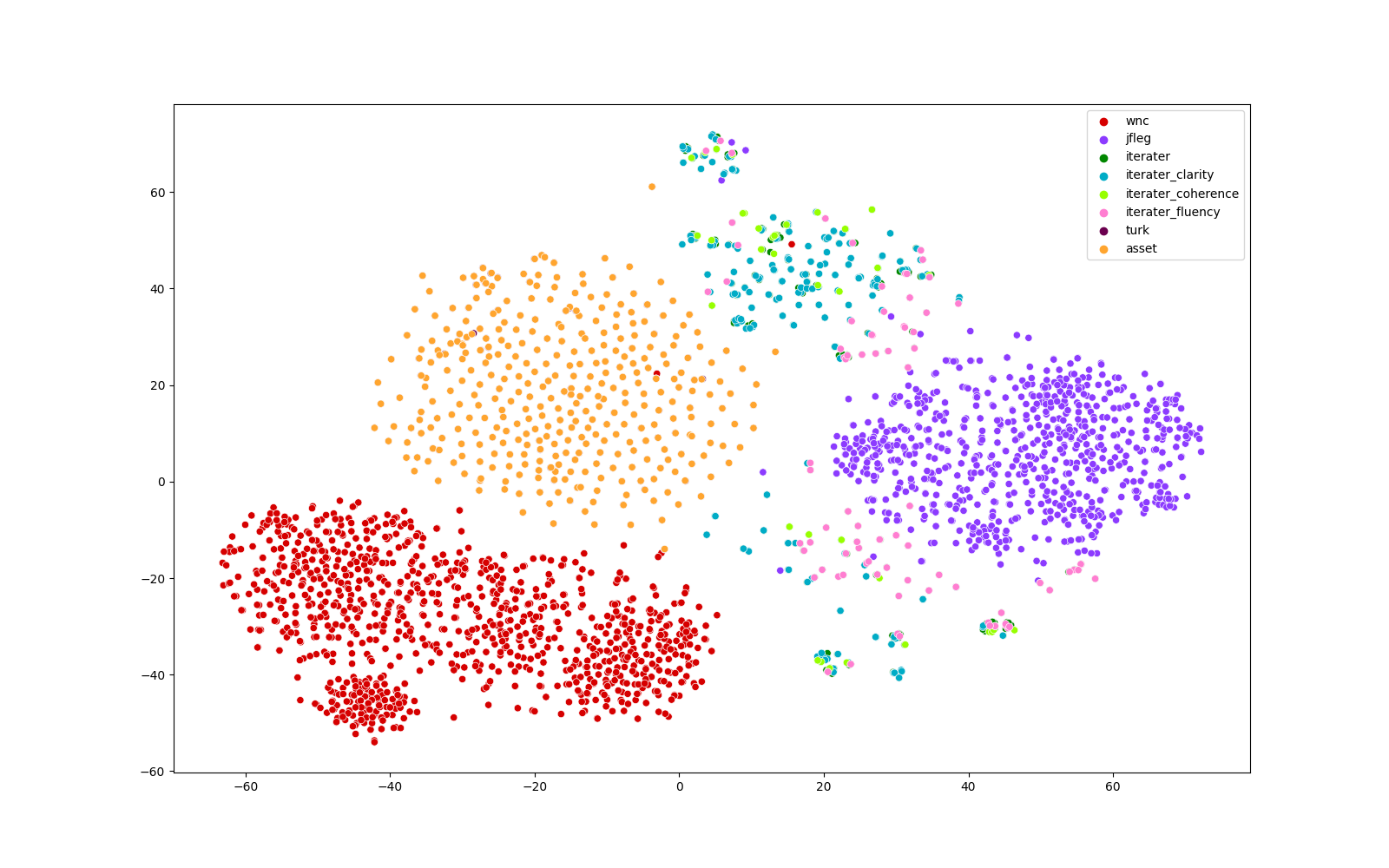}
  \caption{TSNE visualization of the source sentences within all evaluation datasets.}
  \label{fig:appendix-TSNE}
\end{figure*}

For all datasets used in our evaluation, we utilize the publicly available test splits from each dataset. To each data sample (source and target pair), we prepend a randomly selected instructional prompt related to the edit task. For example, for all test samples from TurkCorpus, we prepend a randomly selected instructional prompt from the text simplification choices provided in \citet{raheja2023coedit}. In Table \ref{tab:evaluation-dataset-examples} we provide example test data samples from each evaluation dataset. For context, we additionally provide the sizes of the test splits available for each evaluation dataset. The test splits are as follows: TurkCorpus includes 359 test data samples, Asset includes 359, Iterator Coherence includes 36, Iterator Clarity contains 186, Iterator Fluency contains 88, JFLEG contains 748 and WNC contains 1000. Note, we additionally evaluate on a combined Iterator dataset, noted as Iterator Global in Table \ref{tab:eval_datasets}, which includes all test samples from Iterator Coherence, Clarity and Fluency. The motivation of including an Iterator Global evaluation dataset is to understand model performances on a more generic style-editing task \cite{du2022understanding}.
Furthermore, in Figure \ref{fig:appendix-TSNE}, we provide a TSNE visualization of the evaluation datsets, particularly embedding representations of all source sentences using Sentence-T5 \cite{ni2021sentence}. The visualization demonstrates the diversity among the different datasets, and highlight that the evaluation tasks are not all semantically similar. 


\begin{center}
\begin{table*}[t!]
\resizebox{\textwidth}{!}{%
\renewcommand{\arraystretch}{1.0} 
\begin{tabular}{ l | l | p{6cm} | p{6cm}}\toprule
 \textbf{Evaluation Dataset} & \textbf{Edit Task} & \textbf{Input Example} & \textbf{Output Example} \\
\midrule
 \makecell[l]{TurkCorpus \\ \cite{xu2016optimizing}} & Text Simplification & \textit{Make the sentence simple:} The great dark spot is thought to represent a hole in the methane cloud deck of neptune. & The great dark spot is thought to represent a hole in the methane.
  \\ \midrule
 \makecell[l]{Asset \\ \cite{alva2020asset}} &  Simplification & \textit{Simplify this sentence:} She remained in the United States until 1927 when she and her husband returned to France. & She remained in the United States until returning to France with her husband in 1927.
 \\ \midrule
  \makecell[l]{Iterator Coherence \\ \cite{du2022understanding}} & Coherence & \textit{Fix sentence flow:}  Based on the general linguistic structure of humor, in this paper, we propose a novel approach for detecting humor in short texts by using BERT sentence embedding. &  In this paper, we propose a novel approach for detecting humor in short texts by using BERT sentence embedding .
  \\ \midrule
 \makecell[l]{Iterator Clarity \\ \cite{du2022understanding}} & Clarity & \textit{Write a clearer version for the sentence:}  Using our human-evaluation datasets, we show that existing metrics based on n-gram similarity  do not correlate with human judgments. & Using our human-evaluation datasets, we show that widely used n-gram similarity do not correlate with human judgments.
 \\ \midrule
  \makecell[l]{Iterator Fluency \\ \cite{du2022understanding}} & Fluency & \textit{Fix disfluencies in the sentence:}  In addition, we provide the first robust corpus  this kind for the Brazilian Portuguese language. & In addition, we provide the first robust corpus of this kind for the Brazilian Portuguese language.
 \\ \midrule
 \makecell[l]{JFLEG \\ \cite{napoles2017jfleg}} & Grammar Correction & \textit{Fix the grammar mistakes:} Every person needs to know a bit about math, sciences, arts, literature and history in order to stand out in society. & Every person needs to know a bit about math, science, art, literature and history in order to stand out in society.
 \\ \midrule
  \makecell[l]{WNC \\ \cite{pryzant2020automatically}} & Neutralization & \textit{Remove points of view:} During the unnecessary horseplay, Hamlin fell and severely injured his hand. & During the horseplay, Hamlin fell and severely injured his hand.
  \\ 
 \bottomrule
\end{tabular}}
\caption{Example data samples for each evaluation dataset. Note, the instructional prompt (italicized) are randomly sampled from a list of instructional prompts available in \citet{raheja2023coedit}.}
\label{tab:evaluation-dataset-examples}
\end{table*}
\end{center}

\section{Additional DEFT-UCS Results}
\subsection{Extended Best DEFT-UCS Analysis}
\label{sec:appendix-full-best-deft}
\begin{figure*}[t]
\centering
\begin{subfigure}[b]{0.7\textwidth}
  \centering
  \includegraphics[width=\textwidth]{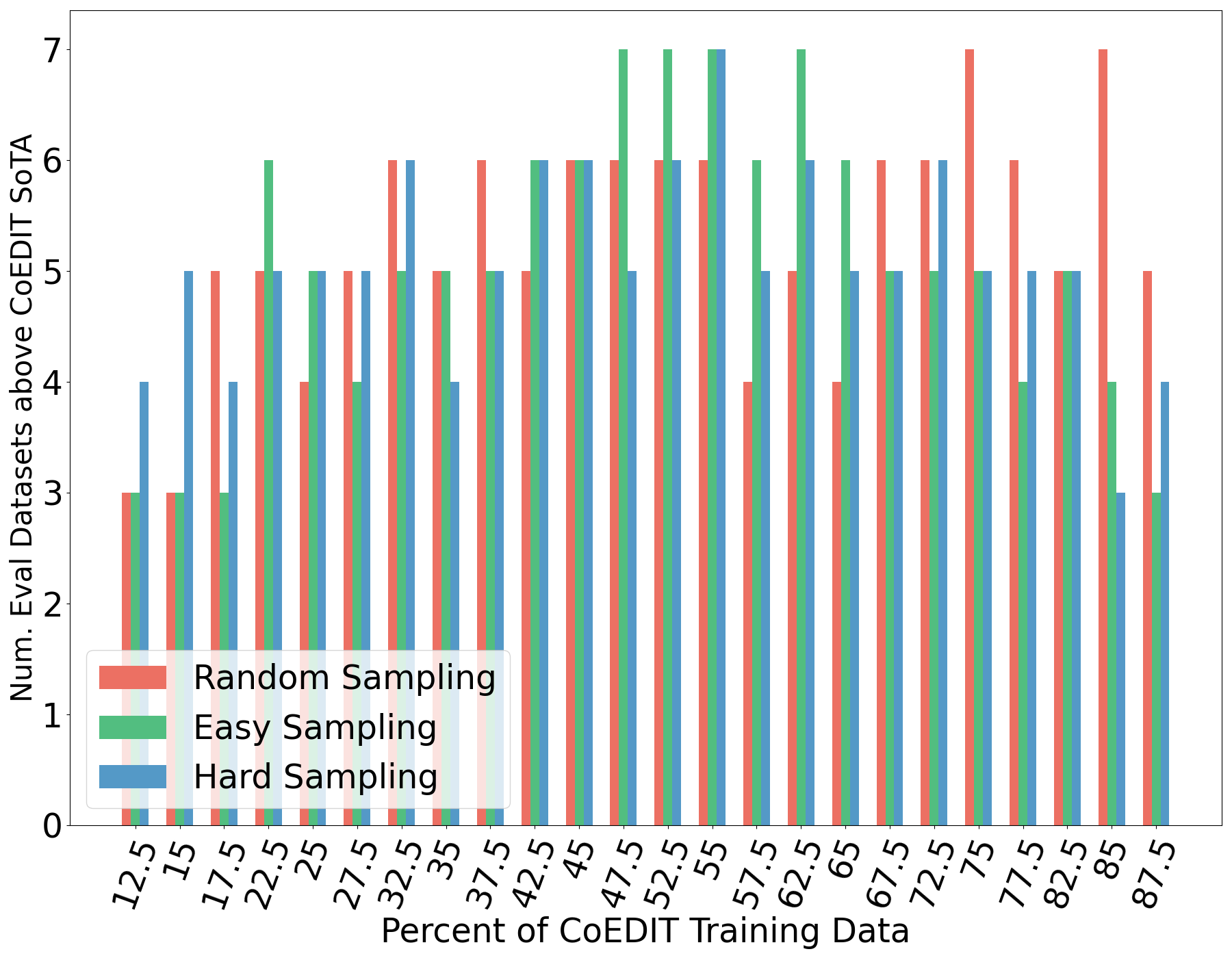}
  \caption[]%
  {{}}
\end{subfigure}\quad
\begin{subfigure}[b]{0.7\textwidth}
  \centering
  \includegraphics[width=\textwidth]{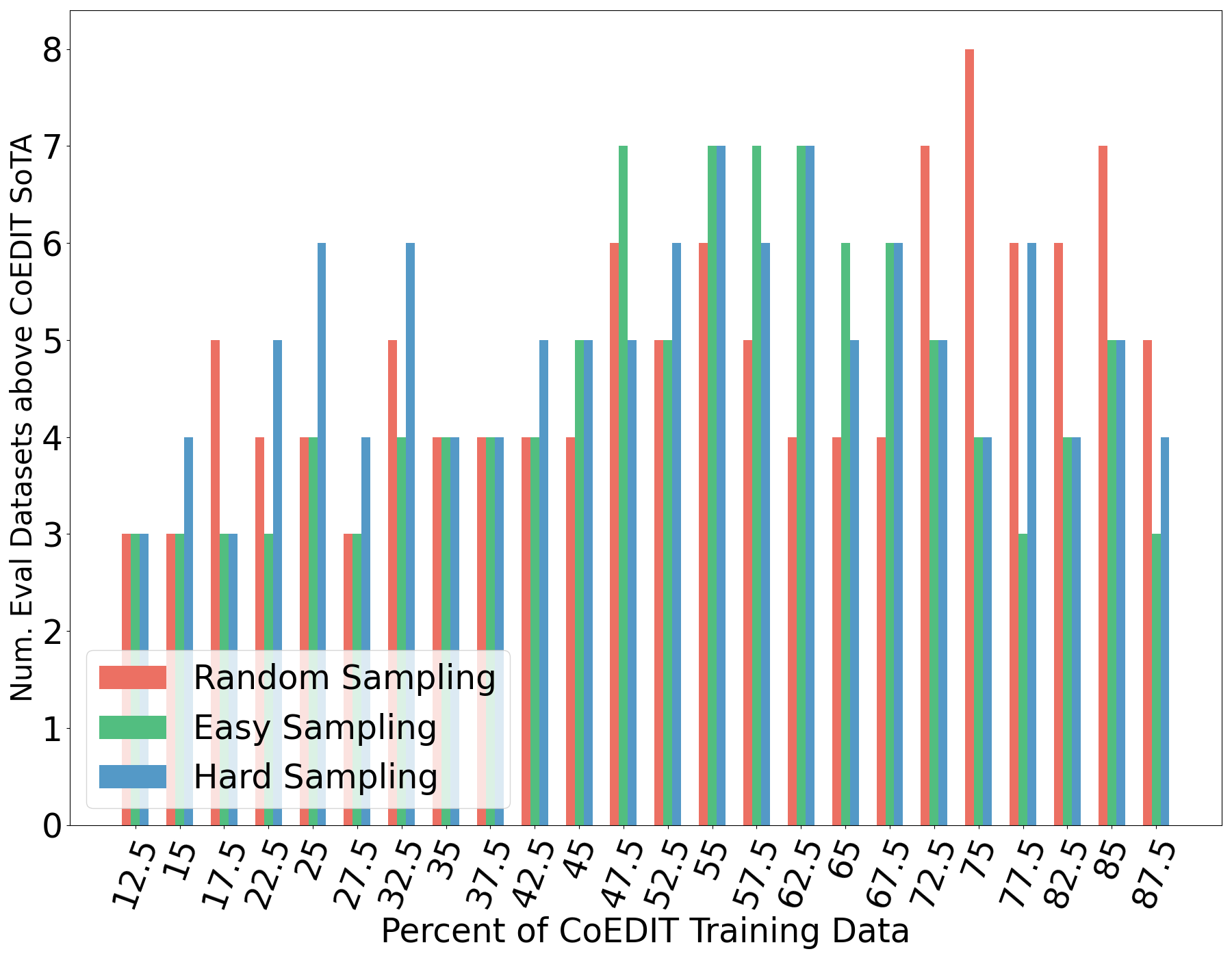}
    \caption[]%
     {{}}
     \end{subfigure}
\caption{Presenting a full analysis, utilizing up to 87.5\% of $D_{CoEDIT}$, on the different sampling methods and amounts of data needed to beat soTA $M_{CoEDIT}$ performance, considering SARI (a) and ROUGE-L (b) scores.}
\label{fig:appendix-full_best_deft}
\end{figure*}

In Section \ref{sec:best-deft-results}, we demonstrate that utilizing on 32.5\% of $D_{CoEDIT}$ can result in an overall best DEFT-UCS model that surpasses $M_{CoEDIT}$ \cite{raheja2023coedit} SARI and ROUGE-L scores on 6 of the 8 evaluation datasets. While Figure \ref{fig:best_deft} in the main paper provides an analysis using up to 45\% of $D_{CoEDIT}$, in this section, we include Figure \ref{fig:appendix-full_best_deft} which provides an exhaustive analysis using up to 87.5\% of $D_{CoEDIT}$. From Figure \ref{fig:appendix-full_best_deft}, we observe that to surpass SARI and ROUGE-L scores on 7 out of the 8 evaluation datasets, 47.5\% of $D_{CoEDIT}$ is necessary. Additionally, we observe that while 75\% of $D_{CoEDIT}$ can be leveraged to surpass ROUGE-L scores on all evaluation datasets. Overall, these results indicate a trade-off between marginal improvement in model performance and the amount of additional data required. 

\subsection{Additional Qualitative Analysis}
\label{sec:appendix-qualitative-examples}
In Table \ref{tab:model-output-examples}, we present example model outputs, qualitatively comparing $M_{CoEDIT}$, $M^{Flan-T5-LG}_{DEFT-UCS}$, $M_{LLAMA-7B}$ and $M_{BLOOM-560M}$. Overall, we observe that the example sentences generated by $M^{Flan-T5-LG}_{DEFT-UCS}$ and $M_{CoEDIT}$ are either identical or similarly edit the input sentence to reflect the edit instruction. When we examine the zero-shot inference outputs from $M_{LLAMA-7B}$ and $M_{BLOOM-560M}$ we observe that these models are not able to produce accurately edited sentences. Instead, we notice repeated generation from both $M_{LLAMA-7B}$ and $M_{BLOOM-560M}$ as well as additional generations that are tangential. These repeated, longer, and irrelevant generated sentences also explain the much lower ROUGE-L observed in Table \ref{tab:best_deft_baselines} within the main paper. Overall, these generated outputs from each model provide further understanding of the need for instruction-tuned LLMs for tasks such as text-editing. These generated output examples also re-iterate that our DEFT-UCS model, $M^{Flan-T5-LG}_{DEFT-UCS}$, can generate similarly edited sentences to the CoEDIT baseline, $M_{CoEDIT}$, while being fine-tuned on ~70\% less data.

\begin{center}
\begin{table*}[t!]
\resizebox{\textwidth}{!}{%
\renewcommand{\arraystretch}{1.0} 
\begin{tabular}{ p{4cm} | p{4cm} | p{4cm} | p{4cm} | p{4cm} }
    \toprule
    & \multicolumn{4}{c}{Generated Model Outputs} \\ \midrule
    \textbf{Input} & \textbf{$M_{CoEDIT}$} & \textbf{$M^{Flan-T5-LG}_{DEFT-UCS}$} & \textbf{$M_{LLaMA2-7B}$} &  \textbf{$M_{BLOOM-560M}$} 
    \\ \midrule
    
    \textit{Make this easier to understand:} the term ``protein'' itself was coined by berzelius, after mulder observed that all proteins seemed to have the same empirical formula and might be composed of a single type of -lrb- very large -rrb- molecule. & The term ``protein'' itself was created by berzelius in 1879, after mulder observed that all proteins seemed to have the same empirical formula and might be composed of a single type of very large molecule. & berzelius thought that all proteins seemed to have the same empirical formula and might be composed of a single type of molecule. & It is a protein made by bacteria that is used to make a large number of other proteins. This is a very large protein that is used to make a large number of other proteins. This is a very large protein that is used to make a large number of other proteins. It is a protein made by bacteria that is used to make a large number of other proteins. It is a protein made by bacteria that is used to make a large number of other proteins. &  The term was coined by mulder to describe the fact that the protein was composed of a single type of -lrb- molecule. The term was coined by mulder to describe the fact that the protein was composed of a single type of -lrb- molecule. The term was coined by mulder to describe the fact that the protein was composed of a single type of -lrb- molecule. 
    \\ \midrule
    \textit{Remove non-neutral POVs:} jimi hendrix (musician), great musician and vocalist died while at girlfriend's monika dannemann hotel room in london. & jimi hendrix (musician), musician and vocalist died while at girlfriend's monika dannemann hotel room in london. & jimi hendrix (musician), musician and vocalist died while at girlfriend's monika dannemann hotel room in london.    & This was an interesting read. I think I'm going to have to edit that article. I've been doing some research on this, and it seems that the story is that Hendrix was at the hotel with Monika Dannemann. She was a German woman who had been a friend of his for some time. She was a model and an artist, and she had been a fan of his music since she was a teenager. She had been in London with him for a couple of weeks, and she had been helping him out with his music. &  the singer was in a coma for a week. the singer was in a coma for a week. the singer was in a coma for a week. the singer was in a coma for a week. the singer was in a coma for a week. the singer was in a coma for a week. the singer was in a coma for a week. the singer was in a coma for a week. the singer was in a coma for a week. the singer was in a coma for a week.
    \\ \midrule
    \textit{Fix grammaticality in this sentence:} The important thing about this case is , each countrey hase to use official solution for their learning problems. & The important thing about this case is that each country has to use an official solution for their learning problems. & The important thing about this case is that each country has to use an official solution for their learning problems. & The important thing about this case is that each country has to use its own official solution for their learning problems.
The important thing about this case is that each country has to use its own official solution for their learning problems. & The main purpose of this paper is to study the grammaticality of the sentence: The important thing about this case is , each countrey hase to use official solution for their learning problems. The main purpose of this paper is to study the grammaticality of the sentence: The important thing about this case is, each countrey hase to use official solution for their learning problems. 

    \\ \midrule
    \textit{Fix disfluencies in the sentence:} To work hard means to run the company. & To work hard means to run a company. & To work hard means to run a company. & Disfluencies are speech disruptions that are often caused by an inability to find the right word. Disfluencies are common in everyday speech. Examples of disfluencies in English 1. To work hard means to run the company. 2. To work hard means to run the company &  The sentence is not a sentence, but a sentence with a disfluency. The sentence is not a sentence, but a sentence with a disfluency. The sentence is not a sentence, but a sentence with a disfluency. The sentence is not a sentence, but a sentence with a disfluency. The sentence is not a sentence, but a sentence with a disfluency.   \\
    \bottomrule
\end{tabular}}
\caption{Example generated, edited sentences from each model for a given input. We observe that non-instruction tuned LMs such as BLOOM-560M and LLAMA-7B mostly struggle in zero-shot inference as demonstrated by the repeated or irrelevant generation.  \citet{raheja2023coedit}.}
\label{tab:model-output-examples}
\end{table*}
\end{center}

\end{document}